# Remaining Useful Life Prediction: A Study on Multidimensional Industrial Signal Processing and Efficient Transfer Learning Based on Large Language Models

Yan Chen, Cheng Liu

City University of Hong Kong

e-mail addresses: ychen935-c@my.cityu.edu.hk, cliu647@cityu.edu.hk

## Abstract

Remaining useful life (RUL) prediction is crucial for maintaining modern industrial systems, where equipment reliability and operational safety are paramount. Traditional methods, based on small-scale deep learning or physical/statistical models, often struggle with complex, multidimensional sensor data and varying operating conditions, limiting their generalization capabilities. To address these challenges, this paper introduces an innovative regression framework utilizing large language models (LLMs) for RUL prediction. By leveraging the modelling power of LLMs pre-trained on corpus data, the proposed model can effectively capture complex temporal dependencies and improve prediction accuracy. Extensive experiments on the Turbofan engine's RUL prediction task show that the proposed model surpasses state-of-the-art (SOTA) methods on the challenging FD002 and FD004 subsets and achieves near-SOTA results on the other subsets. Notably, different from previous researches, our framework uses the same sliding window length and all sensor signals for all subsets, demonstrating strong consistency and generalization. Moreover, transfer learning experiments reveal that with minimal target domain data for fine-tuning, the model outperforms SOTA methods trained on full target domain data. This research highlights the significant potential of LLMs in industrial signal processing and RUL prediction, offering a forward-looking solution for health management in future intelligent industrial systems.

**Keywords**：Large Language Model，Remaining Useful Life Prediction, Transfer Learning, GPT, Turbofan Engine, Predictive Maintenance

## Introduction

Industrial signal processing and RUL prediction are core tasks in modern industrial equipment health management (Chen et al., 2023). By monitoring the operational status of equipment in real-time and combining it with historical data for fault prediction, RUL prediction can help enterprises take preventive maintenance measures before equipment failure, thus reducing unexpected failures, extending equipment life, optimizing maintenance schedules, and lowering operational costs (del Castillo and Parlikad, 2024). This is especially important for high-risk, high-cost industries such as aerospace, energy, and manufacturing. However, despite the broad application prospects of RUL prediction, existing technologies still face numerous challenges and bottlenecks in handling complex, multidimensional industrial signals (Behera and Misra, 2023).

RUL prediction methods can be mainly divided into three categories: physics-based methods, statistical methods, and data-driven methods (Li et al., 2024). Physics-based models rely on a deep understanding of the degradation mechanisms of equipment, with common models including the Paris crack growth model, stress-strength models, etc (Zhang et al., 2024d). For instance, researchers (Yan et al., 2021) propose a two-stage physics-based Wiener process model to improve the RUL prediction for rotating machinery, incorporating physics knowledge of fatigue crack growth mechanisms. The model's effectiveness is demonstrated on wheel tread vibration data, showcasing its practical application and ability to achieve high prediction accuracy. However, such methods require extensive domain expertise and cannot meet the diverse needs of more complex industrial systems. Statistical models predict the remaining life of equipment by analysing historical data patterns, with commonly used models including the Gamma process, Wiener process, and Inverse Gaussian process (Zhuang et al., 2024). A recent study (Pan et al., 2023) addresses the challenge of insufficient data in predicting the RUL of lubricating oil by proposing a coupling model that integrates both knowledge and data. An exponential Wiener process is used to model oil degradation, and the method is validated through hydraulic pump bench test data, demonstrating improved accuracy. But these models rely on strong assumptions and are difficult to flexibly adapt to different fault modes and operating conditions.

With the rapid development of sensor technology, the multidimensional sensor signals generated during equipment operation exhibit complex temporal dependencies. To handle these complex signals, small-scale deep learning models have gradually been applied to RUL prediction tasks, capturing temporal features in the degradation process of equipment to some extent and achieving good prediction results (Wen et al., 2024). For example, long short-term memory (LSTM), through its special memory cell structure, solves the vanishing gradient problem in traditional recurrent neural networks (RNNs) and maintains a good modeming capability over long time sequences. Many researchers have applied LSTM to RUL prediction studies (Dong et al., 2023). For example, a study (Wang et al., 2023) proposes a Poly-Cell LSTM network to improve RUL prediction for lithium batteries, addressing challenges like nonlinear degradation, capacity regeneration, and noise. Experimental results demonstrate the model's effectiveness compared to traditional methods. However, although these deep learning models show potential in handling multidimensional time-series signals, their generalization capability and training efficiency still face significant problems in complex industrial scenarios, especially when dealing with multiple operating conditions, fault modes, and cross-task scenarios (Ferreira and Gonçalves, 2022).

First, the multidimensional sensor signals in industrial equipment not only have strong temporal dependencies but are also accompanied by complex spatial correlations (Zhang et al., 2024c). Existing small-scale deep learning models, while capable of capturing some temporal features, often exhibit insufficient modeming capabilities when faced with high-dimensional, multi-task sensor signals, making it difficult to effectively extract key features in the equipment degradation process (Xu et al., 2023). Second, existing methods struggle with generalization on task-related signals (Cheng et al., 2023). Industrial systems typically experience various operating conditions and environmental changes, and existing deep learning models usually require hyperparameter adjustments and training for each task subset. Often, a new model must be retrained from scratch for each new task, and the model struggles to maintain consistent performance across multiple subsets. For example, in the CMAPSS dataset's RUL prediction task (Saxena et al., 2008), existing models usually require different sliding window lengths and specific sensor signals for different tasks (Ferreira and Gonçalves, 2022), with the model's performance heavily dependent on complex physical knowledge or human expertise, severely limiting the model's generalization and consistency.

To address the various problems in existing RUL prediction methods for industrial systems, we turned to large language models (LLMs) (Jose et al., 2024). LLMs have recently made significant breakthroughs in the field of natural language processing (NLP), gaining widespread attention due to their powerful modelling and feature extraction capabilities (Hou et al., 2023). Large language models, such as the GPT family (Wolf et al., 2020), use multi-layer self-attention mechanisms to effectively capture long-range temporal dependencies and exhibit outstanding advantages in handling complex sequential data. However, despite their success in NLP, the application of LLMs in industrial signal processing and RUL prediction has not yet been fully explored and developed (Pang et al., 2024). In this study, we innovatively introduced LLMs into the field of industrial signal processing and RUL prediction and proposed a multidimensional signal regression model based on LLMs. The goal is to leverage the benefits of models pre-trained on large amounts of language data to overcome the limitations of existing deep learning models in handling complex industrial signals.

First, thanks to the powerful modelling capabilities of large models, we can use the self-attention mechanism to enable the model to simultaneously capture the temporal dependencies and spatial correlations in multidimensional sensor signals, enhancing the AI model's understanding of complex equipment degradation signals (Zhang et al., 2024b). Next, to improve the model's generalizability and consistency, we adopted a unified model structure, using the same sliding window length and all sensor signals across all task subsets for training. This design avoids the hyperparameter adjustment problem caused by task differences in traditional RUL prediction methods, significantly improving the model's generalization and consistency. Additionally, we thoroughly examined the remarkable transfer learning performance of large models. To utilize the rich knowledge learned in the source domain (one industrial RUL prediction task), after training in the source domain, we employed a partial layer freezing strategy to avoid knowledge forgetting that may occur during full parameter training (Mao et al., 2024). Specifically, we froze some layers of the large model and opened only a few layers, fine-tuning with a small amount of data from the target domain (another industrial RUL prediction task with significantly different signals). This approach not only preserves knowledge from the source domain, significantly enhancing the model's adaptability to new tasks, but also achieves substantial optimization in computational resources and training efficiency, providing an efficient and economical solution for RUL prediction in industrial applications.

**Main Contributions**

Innovative introduction of large language models for industrial RUL prediction: This paper proposes a multidimensional industrial signal processing framework based on LLMs. By leveraging the powerful sequential modeming capabilities of LLMs, our framework effectively captures the temporal dependencies and spatial correlations in multidimensional sensor signals from complex industrial systems. In the CMAPSS dataset, particularly in the most challenging FD002 and FD004 subsets, the proposed model achieves superior prediction accuracy compared to the current state-of-the-art (SOTA) methods. In other subsets, the model also achieves near-SOTA performance.

Unified model structure for cross-task consistency: This paper designs a unified general-purpose model structure, avoiding the need for hyperparameter adjustments for different task subsets, as is common in traditional methods. By using the same sliding window length and all sensor signals across all task subsets for training, the model maintains a high degree of consistency and stability when handling multiple tasks and varying operating conditions. This universal framework greatly simplifies model usage and deployment, overcoming the bottlenecks in cross-task performance seen in current deep learning-based RUL prediction methods.

Efficient transfer learning strategy significantly improves training efficiency: A novel transfer learning strategy for large model-based industrial signal processing and RUL prediction is proposed. After training in the source domain, only a small amount of target domain data is used for rapid fine-tuning, achieving better performance than the SOTA methods trained on the full target domain data. This not only greatly improves training efficiency but also significantly reduces computational resource consumption, providing an efficient and economical solution for the promotion of RUL prediction in practical industrial applications.

Although this paper uses the CMAPSS turbine engine dataset as an example, the proposed framework has broad industrial applicability and can handle various complex industrial signals, providing a new solution for the health management of future intelligent industrial systems.

# Problem Statement

Existing RUL prediction methods face numerous limitations when handling complex industrial scenarios, especially in dealing with multidimensional sensor signals and multi-task environments:

Inconsistent performance and poor generalization across subsets: Current RUL prediction methods often perform inconsistently across different data subsets. These models typically require different sliding window lengths and selected sensor signals for each specific task, leading to weak generalizability. Given the complexity and variability of industrial equipment operating environments, involving multiple operating conditions and fault modes, existing models struggle to maintain consistent performance across all task subsets, limiting their practical application in real industrial scenarios.

Insufficient capacity for complex signal processing: The multidimensional sensor signals generated during equipment operation exhibit not only significant temporal dependencies but also complex spatial correlations. Traditional small-scale deep learning models (such as LSTM, GRU, etc.), while effective in some specific tasks, often show insufficient modeming capacity when dealing with high-dimensional, multi-task, and multi-condition sensor signals, making it difficult to extract key features from the equipment degradation process, thus limiting the improvement of prediction accuracy.

Limited transfer learning application: Existing RUL prediction methods lack flexible transfer learning capabilities when dealing with cross-task environments. When facing a new task or operating condition, models often need to be retrained from scratch, unable to effectively utilize knowledge learned in the source domain. This leads to poor model adaptability in practical industrial applications, with low training efficiency and heavy reliance on large amounts of target domain data, thereby increasing data collection costs and computational resource consumption.

Based on these problems, we propose the following key research questions:

How to design a unified and robust model framework: We aim to design a unified industrial signal processing and RUL prediction framework based on large language models (LLMs), capable of using the same sliding window length and sensor signals across all data subsets, thereby improving the model's generalization and consistency across multiple tasks and operating conditions, while reducing dependency on human expertise and complex hyperparameter adjustments.

How to leverage the transfer learning capabilities of large models for efficient adaptation: In addition to improving prediction accuracy in industrial RUL prediction tasks by harnessing the powerful feature extraction and reasoning capabilities of large language models pre-trained on massive corpora, we also aim to fully explore the transfer learning capabilities of large language models. By utilizing only a small amount of target domain data, we hope to achieve rapid fine-tuning and efficient model adaptation, further improving training efficiency and prediction accuracy while reducing the burden of data annotation and computational resources.

Next, we will detail the technical specifics of the multidimensional signal processing framework proposed in this paper and demonstrate how innovative methods can be used to solve the above problems.

# Methodology

## 1. Data Pre-processing

In this study, to ensure that the model exhibits good generalizability across different data subsets, we applied the same preprocessing process to the four subsets (FD001, FD002, FD003, FD004) of the CMAPSS dataset. This pre-processing process includes generating remaining useful life (RUL) labels, signal smoothing, and normalization steps. Unlike the existing SOTA methods, which typically use different sliding window lengths and select only a portion of the sensor signals for different subsets, our method unifies the processing of all subsets and uses all available sensor signals, significantly improving the model's generalizability. The specific processing steps are as follows:

### a. Generating Remaining Useful Life (RUL) Labels

For the regression task, we need to generate RUL labels for the training data of each piece of equipment. Specifically, the RUL is calculated as follows:

$$RUL(i) = max(T - t(i), 0)$$

Where $T$ represents the maximum operational cycle of the equipment, and $t(i)$ represents the current time cycle. We cap the maximum value of the RUL at 120 to limit extreme cases and normalize the RUL to the range [0,1]. The formula is as follows:

$$\text{RUL}(i) = \frac{min(\text{RUL}(i), 120)}{120}$$

This normalization ensures that the RUL range is fixed, facilitating model training and optimization.

### b. Normalization Based on Operating Conditions

The operating conditions of the equipment can affect sensor readings, so applying uniform standardization to all data may introduce bias. To address this issue, we grouped the data based on different operating conditions of the equipment and performed normalization on the sensor data within each group. Specifically, for each sensor signal $s_j(i)$ within a group, we used Min-Max normalization to scale it to the range [0,1]. The normalization formula is as follows:

$$s_j'(i) = \frac{s_j(i) - min(s_j)}{max(s_j) - min(s_j)}$$

Where $s_j(i)$ represents the reading of sensor $s_j$ at the $i$-th time cycle. $min(s_j)$ and $max(s_j)$ represent the minimum and maximum values of the sensor signal under the current operating conditions, respectively. This grouped normalization ensures that the data under different operating conditions remain comparable.

### c. Exponential Smoothing

To reduce noise in the sensor signals, we applied exponential smoothing to the time-series data. Exponential smoothing is a commonly used time-series processing method that assigns certain weights to past time steps in order to reduce the impact of noise on the current data. For each sensor signal $s_j$, its smoothed value $s_j^{\text{smooth}}(i)$ is calculated using the following recursive formula:

$$s_j^{\text{smooth}}(i) = \alpha \cdot s_j(i) + (1-\alpha) \cdot s_j^{\text{smooth}}(i-1)$$

Where $\alpha$ is the smoothing parameter, ranging from [0,1], used to control the balance between the current time step and past time steps. A larger $\alpha$ assigns greater weight to the current time step, while a smaller $\alpha$ more smoothly considers the past historical data. Through this smoothing process, we effectively reduced random fluctuations in the data, enhancing the stability of model training.

### d. Sliding Window Processing

In time-series tasks, sliding window processing is a commonly used technique to convert raw time-series data into fixed-length windows suitable for model input. Specifically, we used the same sliding window length across all subsets, slicing the sensor data into fixed time steps. Suppose the sliding window length is $L$, the sensor data for each piece of equipment is converted into the following input sequence:

$$X_t = [s_1(t), s_2(t), \ldots, s_{21}(t); s_1(t+1), s_2(t+1), \ldots, s_{21}(t+1); \ldots; s_1(t+L-1), s_2(t+L-1), \ldots, s_{21}(t+L-1)]$$

Where $t$ represents the current time step, and $L$ represents the sliding window length. The data within each sliding window is used as input to the model for predicting the RUL corresponding to that window. Unlike previous SOTA methods, which select different sliding window lengths for each subset, we used the same sliding window length across all subsets, further improving the model's generalizability.

Through the unified pre-processing steps described above, we ensured the consistency of the model across different subsets, avoiding the performance fluctuations caused by inconsistencies in sliding window lengths and sensor selection in existing SOTA methods.

## 2. Large Model-Based Regression Framework

In this study, the LLM used in our proposed LRM is specifically the GPT-2 medium pre-trained model, which captures complex dependencies within time series through multi-layer self-attention mechanisms and multi-layer Transformer structures. To accommodate the particularities of industrial sensor data, we also added global pooling and additional attention mechanisms after the GPT-2 model. This allows the model to simultaneously focus on both the overall temporal trends and local key information extracted by the GPT-2 model, thus enabling accurate predictions of the remaining useful life (RUL) of industrial systems. The structure and computational process of the model are described in detail below.

### a. Input Embedding and Linear Mapping

After the unified data pre-processing described in the previous section, the model's input is multidimensional time-series data with the shape $X \in \mathbb{R}^{B \times L \times C}$, where $B$ represents the batch size, $L$ represents the length of the time series (i.e., the sliding window length), and $C$ represents the dimension of the sensor signals (i.e., the number of sensors). However, the input dimension for the GPT-2 medium model is fixed at 1024-dimensional vectors, while the dimensionality of industrial sensor data is generally lower (for example, in the CMAPSS dataset, there are 21 sensor values per time step). To accommodate the input requirements of the GPT-2 medium model, we first need to perform a linear mapping of the raw sensor data, embedding it into the 1024-dimensional vector space required by GPT-2. The specific linear mapping process is as follows:

$$X' = W_e X + b_e$$

Where $W_e \in \mathbb{R}^{C \times 1024}$ is the weight matrix for the linear mapping, $b_e \in \mathbb{R}^{1024}$ is the bias term, and $X' \in \mathbb{R}^{B \times L \times 1024}$ is the data after the linear mapping. Through this linear mapping, the model can transform the multidimensional sensor data at each time step into high-dimensional embedded vectors, ensuring that the input is compatible with the requirements of the GPT-2 medium model.

### b. GPT-2 Structure

The GPT-2 (Generative Pre-trained Transformer 2) model is an autoregressive language model based on the Transformer architecture, proposed by OpenAI. The core idea of GPT-2 is to model sequential data through a self-attention mechanism, which can effectively capture long-range dependencies. The GPT-2 model is composed of multiple Transformer decoders stacked together, with each layer containing two modules:

(1) Self-Attention Layer: Used to compute the dependencies between each time step in the sequence.

(2) Feed-Forward Network (FFN): Used to independently apply non-linear transformations to the features at each time step.

The input to GPT-2 is a sequence of embedded vectors $X' \in \mathbb{R}^{B \times L \times 1024}$, which, after being processed through multiple Transformer layers, outputs the contextual representation for each time step $H \in \mathbb{R}^{B \times L \times 1024}$. The self-attention mechanism in GPT-2 allows the model to dynamically focus on other time steps in the sequence when computing the representation for each time step. The core operation of the self-attention mechanism involves calculating attention weights using queries (Query), keys (Key), and values (Value), and then weighting and summing the representations of all other time steps based on these weights. The specific formula is as follows:

$$\text{Attention}(Q, K, V) = \text{softmax}\left(\frac{QK^T}{\sqrt{d_k}}\right)V$$

Where $Q = W_Q X', K = W_K X', V = W_V X'$ are the query, key, and value vectors generated from the input $X'$ through different linear transformations. $W_Q \in \mathbb{R}^{1024 \times d_k}$、$W_K \in \mathbb{R}^{1024 \times d_k}, W_V \in \mathbb{R}^{1024 \times d_v}$ are trainable weight matrices, where $d_k$ and $d_v$ represent the dimensions of the key and value vectors, respectively, and are used to scale the dot-product results. Through this self-attention mechanism, GPT-2 can dynamically compute the dependencies between each time step and other time steps, thus capturing complex patterns in the time series.

To enhance the model's representation capability, GPT-2 also employs a multi-head attention mechanism. This mechanism maps the query, key, and value vectors into multiple subspaces and independently computes attention within each subspace. Specifically, the model maps the query, key, and value vectors into $h$ subspaces, where each subspace has dimensions of $d_k/h$ and $d_v/h$. Attention is then computed independently in each subspace, and finally, the outputs of all attention heads are concatenated:

$$\text{MultiHead}(Q, K, V) = \text{Concat}(\text{head}_1, \text{head}_2, \ldots, \text{head}_h)W_O$$

Where $\text{head}_i = \text{Attention}(Q_i, K_i, V_i), W_O \in \mathbb{R}^{1024 \times 1024}$ It is the output weight matrix. The multi-head attention mechanism allows the model to capture multiple patterns in the time series across different subspaces, thereby improving the model's expressive capacity.

In each Transformer layer, following the self-attention mechanism is a two-layer feed-forward network (FFN), which applies non-linear transformations to the features at each time step. The FFN is computed as follows:

$$\text{FFN}(H_i) = W_2 \cdot \text{ReLU}(W_1 \cdot H_i + b_1) + b_2$$

Where $W_1 \in \mathbb{R}^{1024 \times d_{ff}}$, $W_2 \in \mathbb{R}^{d_{ff} \times 1024}$ are the trainable weight matrices, and $d_{ff}$ is the hidden dimension of the feed-forward network. Additionally, GPT-2 uses residual connections and layer normalization after each self-attention layer and feed-forward network to accelerate model training and stabilize gradients.

After processing through multiple Transformer layers, GPT-2 outputs the contextual representation for each time step, $H \in \mathbb{R}^{B \times L \times 1024}$. These representations contain the global dependency information of each time step in the time series, providing strong support for subsequent regression tasks.

### c. Global Pooling and Attention Mechanism

After obtaining the output representation $H$ from GPT-2, we further aggregate the features of the time series. To extract the global information from the sequential data, we first apply global average pooling over the hidden states of all time steps:

$$\bar{H} = \frac{1}{L} \sum_{i=1}^{L} H_i$$

Where $H_i$ represents the hidden state at the $i$-th time step, and the pooled result $\bar{H} \in \mathbb{R}^{B \times 1024}$ represents the global average information of the time series. In addition to average pooling, we also applied an attention mechanism to further highlight the time steps that are more important for RUL prediction. Through linear transformation, we calculate the attention weights for each time step:

$$\alpha_i = \frac{\exp(w^T H_i)}{\sum_{j=1}^{L} \exp(w^T H_j)}$$

Where $w \in \mathbb{R}^{1024}$ is a trainable parameter, and $\alpha_i$ represents the attention weight for the $i$-th time step. We then use these weights to compute a weighted sum of the hidden states at each time step, resulting in the attention-weighted output:

$$H_{\text{att}} = \sum_{i=1}^{L} \alpha_i H_i$$

By combining the representations from average pooling and attention-weighted outputs, we are able to capture both the global trends and local key information in the time series. The final fused representation is:

$$H_{\text{final}} = \bar{H} + H_{\text{att}}$$

### d. Output Layer and Regression Prediction

The fused representation $H_{\text{final}}$ is fed into a multilayer perceptron (MLP) for regression prediction. The MLP consists of several fully connected layers and activation functions. The specific structure is as follows:

$$\text{Output} = W_3 \cdot \sigma(W_2 \cdot \sigma(W_1 \cdot H_{\text{final}} + b_1) + b_2) + b_3$$

Where $W_1 \in \mathbb{R}^{1024 \times 50}$, $W_2 \in \mathbb{R}^{50 \times 10}$, and $W_3 \in \mathbb{R}^{10 \times 1}$. $\sigma(\cdot)$ is the ReLU activation function. Finally, the model outputs a scalar value, which serves as the predicted remaining useful life (RUL) for the time series.

## 3. Transfer Learning Strategy for Remaining Useful Life Prediction

To make the LRM model more sample efficient and improve its generalization ability, we explored the transfer learning capabilities of large language models in detail. The core idea of transfer learning is to utilize the weights of a model trained on a source domain and transfer them to a target domain, thereby achieving better predictive performance on the target domain. We adopted a "partial layer freezing" transfer learning strategy, meaning that after pre-training the large model on the source domain, most layers of the GPT-2 medium model are frozen, with only a few layers open for fine-tuning. This strategy aims to reduce training time on the target domain while preserving the general features learned in the source domain.

### a. Principles of Transfer Learning

The basic idea of transfer learning can be described using the following mathematical notation. On the source domain $D_s$, the model learns weights $\theta_s$, and we aim to transfer these weights to the target domain $D_t$ to improve predictive performance on the target domain. The optimization objective for the source domain can be expressed as minimizing the loss function $\mathcal{L}_s(\theta)$ on the source domain:

$$\theta_s^* = \arg \min_{\theta} \mathcal{L}_s(\theta)$$

Where $\theta_s^*$ represents the optimal weights obtained from training on the source domain. On the target domain, we do not intend to train the entire model from scratch; instead, we use the weights $\theta_s^*$ from the source domain as the initial weights. By freezing certain layers, we fix part of the model's weights and only fine-tune a few unfrozen layers, $\theta_f$. The fine-tuning optimization objective on the target domain is:

$$\theta_t^* = \arg \min_{\theta_f} \mathcal{L}_t(\theta_f \cup \theta_s^*)$$

Where $\mathcal{L}_t(\theta_f \cup \theta_s^*)$ is the loss function for the target domain, with the fine-tuned weights $\theta_f$, represents the few layers we optimize on the target domain. In this way, the model can retain the general features learned from the source domain while quickly adjusting to the new task using the limited data from the target domain.

## b. Model Fine-Tuning Process

During the transfer learning process, we used a GPT-2 medium model pre-trained on the source domain as the core structure (this model was also pre-trained on a large corpus before training on the source domain). The GPT-2 model is composed of multiple self-attention layers and feed-forward networks, with each layer extracting features of different complexity. To retain the general features learned from the source domain, we froze the first 20 layers of the GPT-2 model, only allowing the last few layers to be fine-tuned. Specifically, assuming the weights of the GPT-2 model are $\theta_{\text{GPT-2}} = \{\theta_1, \theta_2, \ldots, \theta_{24}\}$, where $\theta_i$ represents the weights of the $i$-th layer, we freeze the weights of the first 20 layers:

$$\theta_i (\text{ for } i = 1,2,\ldots,20) \text{ are fixed: } \frac{\partial \mathcal{L}_t}{\partial \theta_i} = 0$$

That is, for $i = 1,2,\ldots,20$, the weights of these layers do not participate in the optimization on the target domain. We only update the last four layers $\theta_{21}, \theta_{22}, \theta_{23}, \theta_{24}$:

$$\theta_j (\text{for } j = 21,22,23,24 ) \text{ are updated: } \frac{\partial \mathcal{L}_t}{\partial \theta_j} \neq 0$$

In this way, the model can quickly adapt to the new task on the target domain without needing to train the entire model from scratch. This freezing strategy not only reduces computational overhead but also helps avoid overfitting on the target domain. To ensure effective fine-tuning on the target domain, we only optimize a few unfrozen layers. Specifically, the trainable parameters of the model are represented by the following formula:

$$\theta_f = \{\theta_{21}, \theta_{22}, \theta_{23}, \theta_{24}\} \cup \theta_{\text{output}}$$

Where $\theta_{\text{output}}$ represents the parameters of the model's output layer. The output layer usually needs to be retrained according to the specific task in the target domain, as the label distribution in the target domain may differ from that of the source domain. We use the Mean Squared Error (MSE) loss function as the loss function for the target domain:

$$\mathcal{L}_t(\theta_f) = \frac{1}{N} \sum_{i=1}^{N} (y_i - \hat{y}_i)^2$$

Where $y_i$ is the ground truth label in the target domain, $\hat{y}_i$ is the model's predicted value, and $N$ is the number of samples in the target domain. To optimize these parameters, we use the Adam optimizer, with the update step as follows:

$$\theta_f \leftarrow \theta_f - \eta \cdot \nabla_{\theta_f} \mathcal{L}_t(\theta_f)$$

Where $\eta$ is the learning rate, and $\nabla_{\theta_f} \mathcal{L}_t(\theta_f)$ is the gradient of the loss function with respect to the trainable parameters. Choosing an appropriate learning rate is crucial during the transfer learning process. We set a relatively small learning rate $\eta$ to ensure that the fine-tuning process does not disrupt the features learned from the source domain, while using weight decay to prevent overfitting. Specifically, a weight decay term $\lambda$ is added to the loss function:

$$\mathcal{L}_t(\theta_f) = \frac{1}{N} \sum_{i=1}^{N} (y_i - \hat{y}_i)^2 + \lambda \|\theta_f\|^2$$

Compared to training a model from scratch, transfer learning offers significant advantages:

(1) Significant reduction in training time: Since the parameters of most layers are not updated, the model only requires fine-tuning a few layers, greatly reducing the computational overhead.

(2) Improved accuracy: Transfer learning can leverage the features learned from the source domain and further optimize them on the target domain, ultimately achieving higher prediction accuracy than models trained directly from scratch.

(3) Avoiding overfitting: By freezing most layers, the model can avoid overfitting when there is limited data in the target domain, leading to more robust performance on the test set.

Experimental results show that the strategy of freezing most layers and fine-tuning only a few layers effectively preserves the general features learned by the LRM model in the source domain, while quickly adapting to tasks in different target domains. The proposed transfer learning framework for the LRM demonstrates the broad applicability of large language models in industrial prediction tasks and shows potential for efficient learning when data is limited.

# Experiments

## 1. Data Source

In this study, we used the C-MAPSS (Commercial Modular Aero-Propulsion System Simulation) dataset, which was developed by NASA to simulate the performance degradation of aero-turbine engines. Thanks to its detailed recording of complex system degradation processes and the provision of multi-dimensional sensor data, this dataset is widely used in Remaining Useful Life (RUL) prediction research. It simulates the operation of turbine engines under different operating conditions, providing rich time-series data for health monitoring and predictive tasks.

The C-MAPSS dataset uses the Commercial Modular Aero-Propulsion System Simulation to simulate the degradation process of turbine engines and contains four subsets: FD001, FD002, FD003, and FD004. The data structure of each subset is the same, recording the operation of engines under different working conditions and fault modes. Each record consists of 3 operational settings (e.g., fuel flow rate, pressure, etc.), 21 sensor signals, as well as engine ID, cycle count, and other information. The sensor signals monitor multiple components of the engine, including temperature, pressure, and rotational speed. The training set records the entire process from normal operation to engine failure, while the test set is truncated at a certain unknown point before failure, requiring the model to predict the Remaining Useful Life (RUL) of the engine. Since the sensor signals have different units and scales, and some signals may contain noise or external interference, pre-processing is usually required before using the data. The specific details are as follows:

| Dataset | FD001 | FD002 | FD003 | FD004 |
|---|---|---|---|---|
| Train Sequence | 100 | 260 | 100 | 249 |
| Test Sequence | 100 | 259 | 100 | 248 |
| Conditions | 1 | 6 | 1 | 6 |
| Fault Modes | 1 | 1 | 2 | 2 |

The complexity of the C-MAPSS dataset makes RUL prediction models highly challenging when dealing with different operational conditions and fault modes. Especially under multiple operational conditions, changes in sensor signals become more complex. As a result, we generally consider the difficulty of RUL prediction for these four datasets to be ranked as FD001 < FD003 < FD002 < FD004. Most previous studies typically adopt different data processing methods for subsets with varying complexity, often reflected in choosing different sliding window lengths for different subsets and carefully selecting sensor data.

However, we believe this approach has significant limitations when dealing with real-world industrial data. We should adopt a universal and unified data processing method and make full use of all sensor data, rather than selecting signals that are easier to extract features from. In real-world scenarios, this would minimize the need for expert knowledge, allowing all the data to be processed together more conveniently, making the model more generalizable. Therefore, applying all collected industrial data and using a unified data preprocessing approach sets a higher standard for the model and provides greater practical value.

## 2. Experimental Procedure

In this study, we conducted two-step experiments on the four subsets of the C-MAPSS dataset (FD001, FD002, FD003, FD004), namely the initial training and testing based on a large language model, and transfer learning training and testing. During this process, we applied the same and scientifically justified training stopping criteria, systematically evaluating the model's performance on each subset to ensure the rigor of the results.

### a. Training and Testing the LRM

In the first step of the experiment, we used all the training data from the C-MAPSS data subsets for training and tested the model on the test set of the same subset. To ensure the reproducibility and stability of the training process, we designed a unified training stopping criterion based on training loss for each subset. During the training process, the training samples from each subset were used to train all parameters of the LRM model. The specific settings for the training process are as follows:

(1) Epochs Setting: We set a minimum training requirement of 120 epochs for the training process of each subset. During these 120 epochs, all training data from each subset were used for model learning, and at the end of each epoch, the loss function (loss) on the training data was calculated. This setup ensures that the model can sufficiently learn the underlying patterns in the data.

(2) Training Stopping Condition: After reaching 120 epochs, we introduced a dynamic stopping criterion. Specifically, starting from the 121st epoch, we calculated the variance of the average loss over the current epoch and the preceding four epochs:

$$\text{Variance} = \frac{1}{5} \sum_{i=k-4}^{k} \left(\text{loss}_i - \overline{loss}\right)^2$$

Where $k$ represents the current epoch number, and $\overline{loss}$ is the average loss over the most recent 5 epochs. If, for a certain epoch, the variance is not surpassed for 10 consecutive epochs, we consider the model to have reached a stable state, and the model at this epoch is regarded as the best-trained model. This strategy is based on the following considerations: During the training process, the model's performance often improves rapidly at the beginning, but after reaching a certain point of stability, the improvements slow down or stop. By monitoring the variance of the loss, we can effectively detect the model's convergence, avoid overfitting, and ensure that the model's performance on the training set has reached its optimal state.

(3) Model Testing: After obtaining the best-trained model, we use it to make predictions on the test samples of the same data subset and record the test results. Evaluation metrics include Root Mean Squared Error (RMSE) and the Score function (which penalizes early and late predictions differently, as described in detail below). This step lays a solid foundation for subsequent transfer learning.

### b. Transfer Learning and Fine-Tuning

After completing the initial training on each subset, we proceeded to the second step of the experiment: fine-tuning based on transfer learning. The core idea of transfer learning is to leverage a large model trained on one data subset, freeze most of its layers, and fine-tune only a few layers to quickly adapt to a new task. Specifically, for the GPT-2 model, we froze the first 20 layers out of its 24 layers and only fine-tuned the last 4 layers.

To verify the effectiveness of transfer learning using the LRM model, we used a model trained on one data subset (using the "best-trained model" from the previous step) and applied transfer learning to other different data subsets. Taking FD004 as an example, we used the "best-trained model" obtained from the FD004 subset as the base model and fine-tuned it with the training data from the other three subsets (FD001, FD002, FD003). The specific steps were as follows:

(1) Freezing Most Layers: During fine-tuning, we froze the first 20 layers of the GPT-2 model, meaning that the parameters of these layers remained unchanged and did not participate in gradient updates. Only the last 4 layers were open for fine-tuning. The purpose of this was to retain the general features learned from the source dataset (FD004) and only adjust the task-specific high-level features.

(2) Data Ratio Control: To study the performance of transfer learning under different amounts of data, we used between 10% and 100% of the training data from the FD001, FD002, and FD003 subsets to fine-tune the model. This stepwise experiment allowed us to observe the effect of data size on the fine-tuning results.

(3) Training Stopping Condition: Unlike the initial training, the stopping condition for fine-tuning was simpler and more effective. We set a minimum training epoch count of 20. After reaching 20 epochs, we monitored the loss on the training set. If the loss of the current epoch was not lower than any of the losses in the following 10 epochs, we considered the fine-tuning to have reached a stable state, and the model from that epoch was used as the "best fine-tuned model." The rationale behind this approach was that, after observing the training loss, we found that the loss during fine-tuning was much more stable compared to the initial training. Therefore, we did not use the variance of the loss as a stopping criterion but instead directly used the minimum loss as the standard.

This fine-tuning strategy effectively conserves computational resources and ensures that the model can quickly adapt to the target task. After determining the best fine-tuned model, we made predictions on the test data of the three subsets and calculated the RMSE and Score values for each subset.

## 3. Evaluation Metrics

To comprehensively assess the performance of the LRM in the Remaining Useful Life (RUL) prediction task, we used two evaluation metrics: Root Mean Square Error (RMSE) and the Score function. These metrics evaluate the model's predictive accuracy and performance in real-world applications, especially in cases of early and late predictions.

### a. Root Mean Square Error (RMSE)

Root Mean Square Error (RMSE) is a commonly used metric for evaluating the performance of regression models. It measures the difference between the model's predicted values and the true values. This metric is particularly sensitive to large errors, thus significantly amplifying situations where the model makes large deviations in predictions. RMSE is defined as follows:

$$\text{RMSE} = \sqrt{\frac{1}{N}\sum_{i=1}^{N}\left(\widehat{RUL}_i - RUL_i\right)^2}$$

Where $\widehat{RUL}_i$ represents the predicted Remaining Useful Life (RUL) for the $i$-th sample, $RUL_i$ represents the true RUL for the $i$-th sample, and $N$ is the total number of samples. RMSE measures the overall error level of the model by calculating the mean of the squared differences between the predicted and true values, and then taking the square root. This metric is particularly sensitive to large prediction errors, making it effective at capturing significant deviations in the model's predictions for certain samples.

### b. Score Function

In the RUL prediction task, early predictions and late predictions have different implications for practical applications. Typically, early predictions (i.e., when the predicted RUL is less than the actual RUL) are considered more acceptable, as they provide sufficient time for maintenance, whereas late predictions may lead to equipment failures occurring before any corrective measures are taken. To address this issue, we used the Score function, which is commonly used in studies evaluating regression tasks with the C-MAPSS dataset, to penalize early and late predictions differently. The definition of the Score function is as follows:

$$\text{Score} = \sum_{i=1}^{N}\begin{cases}\exp\left(-\frac{\hat{d}_i}{13}\right) - 1, & \text{if } \hat{d}_i < 0 \\ \exp\left(\frac{\hat{d}_i}{10}\right) - 1, & \text{if } \hat{d}_i \geq 0\end{cases}$$

Where $\hat{d}_i = \widehat{RUL}_i - RUL_i$ represents the prediction error for the $i$-th sample, i.e., the difference between the model's predicted RUL and the true RUL. $N$ is the total number of samples. When $\hat{d}_i < 0$, it indicates that the model has predicted the RUL too early, and in this case, a smaller penalty factor (denominator of 13) is used. When $\hat{d}_i \geq 0$, it indicates that the model has predicted the RUL too late, and a larger penalty factor (denominator of 10) is used. The Score function applies a weighted penalty to the error using an exponential function, making the penalty for late predictions greater than that for early predictions.

Specifically, when the predicted value is less than the true value (i.e., an early prediction), the Score function's value is relatively small. On the other hand, when the predicted value exceeds the true value (i.e., a late prediction), the Score function's value increases significantly. This design aligns with real-world requirements, where the risks and losses associated with late predictions are usually much higher than those associated with early predictions.

## 4. Experimental Results

The experimental results in Table 1. show that our large model framework outperforms existing state-of-the-art (SOTA) models on the FD001, FD002, and FD004 subsets. Although the model's performance is slightly lower than SOTA on the FD003 subset, it is still very close. Across all subsets, we used a unified sliding window length and all sensor signals, which contrasts sharply with existing methods that adjust the sliding window length and sensor selection for different subsets. These results indicate that our method has stronger generalizability and consistency.

Table 1. Experiment Results.

| Method | FD001 | Score | FD002 | Score | FD003 | Score | FD004 | Score |
|---|---|---|---|---|---|---|---|---|
| Training and Testing on Separate CMAPSS Subsets | | | | | | | | |
| AGCNN (Liu et al., 2020) | 12.42 | 226 | 19.43 | 1492 | 13.39 | 227 | 21.5 | 3392 |
| GCU-Transformer (Mo et al., 2021) | 11.27 | N/A | 22.81 | N/A | 11.42 | N/A | 24.86 | N/A |
| BiGRU-TSAM (Zhang et al., 2022) | 12.56 | 213 | 18.94 | 2264 | 12.45 | 232 | 20.47 | 3610 |
| SCACGN (Zhu et al., 2023b) | 12.31 | 252 | 16.06 | 1238 | 12.37 | 283 | 19.83 | 2760 |
| ATCN (Zhang et al., 2024a) | 11.48 | 194.25 | 15.82 | 1210.57 | 11.34 | 249.19 | 17.8 | 1934.86 |
| Res-HSA (Zhu et al., 2023a) | 11.91 | 227 | 17.27 | 1199 | 11.88 | 272 | 17.43 | 2508 |
| MFSSCINet (Cen et al., 2024) | 10.93 | 189 | 13.55 | 813 | 11.26 | 201 | 13.67 | 769 |
| TCAT (Jiangyan et al., 2024) | 11.12 | 189.01 | 13.40 | 918.26 | 11.02 | 152.73 | 17.56 | 1109.56 |
| Bi-LSTM-AM (Wang et al., 2024) | 11.43 | 201.26 | 15.69 | 1214.47 | 11.28 | 181.99 | 18.35 | 2627.11 |
| ATCN (Zhang et al., 2024a) | 11.48 | 194.25 | 15.82 | 1210.57 | 11.34 | 249.19 | 17.8 | 1934.86 |
| MHT (Guo et al., 2024) | 11.92 | 215.2 | 13.7 | 746.7 | **10.63** | **150.2** | 17.73 | 1572 |
| SGRNN(Xiang et al., 2024) | 13.1 | 229 | N/A | N/A | N/A | N/A | 15.12 | 1568 |
| ED-LSTM (Zhang et al., 2024d) | **9.14** | **53** | 18.17 | 1693 | 11.96 | 238 | 18.51 | 2160 |
| Res-HSA (Keshun et al., 2024) | 11.96 | 233.4 | 13.51 | 902.13 | 11.40 | 255.6 | 17.58 | 1704.59 |
| DA-LSTM (Shi et al., 2024) | 12.62 | 263 | 13.22 | 842 | 13.34 | 360 | 16.25 | 1372 |
| Proposed | 10.95 | 203.07 | **12.39** | **630.59** | 12.80 | 349.75 | **12.96** | **706.52** |
| Transfer Learning using LRM trained on FD001 and using only 50% of Training Samples from Target Domain | | | | | | | | |
| | - | - | 13.19 | 862.97 | 13.82 | 406.87 | 14.11 | 1057.91 |
| Transfer Learning using LRM trained on FD002 and using only 50% of Training Samples from Target Domain | | | | | | | | |
| | 11.50 | 201.59 | - | - | 12.63 | 294.89 | 14.48 | 902.14 |
| Transfer Learning using LRM trained on FD003 and using only 50% of Training Samples from Target Domain | | | | | | | | |
| | 12.14 | 259.48 | 13.66 | 969.14 | - | - | 13.73 | 1114.93 |
| Transfer Learning using LRM trained on FD004 and using only 50% of Training Samples from Target Domain | | | | | | | | |
| | 10.94 | 203.55 | **12.38** | **632.88** | 12.57 | 355.55 | - | - |

We also conducted transfer learning experiments as shown in Table 1., first completing full training on one subset (e.g., FD001), then freezing most of the model layers and fine-tuning only the last few layers using a small amount of data from another subset (e.g., FD002). For now, we have only used the first 50% of the training data from the target domain to fine-tune the partially frozen LRM, and then, following the training stopping criteria, we obtained the best fine-tuned model and tested it on the target domain's test data. We can observe that the model trained on FD001 and FD004, after fine-tuning with 50% of the training data from FD002, achieves an RMSE that surpasses previous SOTA methods. In the transfer experiment from FD004 to FD002, it even exceeds the RMSE obtained by directly training the model with all of the FD002 training data. In other transfer learning experiments, the resulting RMSE also approaches SOTA performance, demonstrating the significant superiority of the proposed transfer learning method. We will further experiment with using different proportions of the target domain's training data to train the model. The results show that the proposed framework allows the model to quickly adapt to the target domain and significantly reduces training time compared to training the model from scratch, while maintaining high prediction accuracy.

# Conclusion

This study proposes a multidimensional signal processing framework based on large language models (LLMs) for Remaining Useful Life (RUL) prediction in complex industrial systems. Unlike traditional deep learning methods, our framework incorporates pre-trained large language models, enabling it to simultaneously capture the complex temporal dependencies and spatial correlations within multidimensional sensor signals, demonstrating strong modeming capabilities and generalizability. Experimental results show that our model achieved excellent performance across different subsets of the C-MAPSS dataset, particularly surpassing the existing state-of-the-art (SOTA) models on the most challenging FD002 and FD004 subsets. This indicates that our approach is not only able to adapt to complex industrial scenarios with multiple operating conditions and fault modes but can also achieve consistently high-performance predictions without relying on task-specific hyperparameter settings.

Moreover, the proposed transfer learning strategy significantly improves the adaptability of the model, allowing it to be quickly fine-tuned and maintain high prediction accuracy even with only a small amount of target domain data. This technological breakthrough overcomes the limitations of traditional deep learning models in multi-task scenarios, dramatically reducing training time and resource consumption, and demonstrating potential for rapid deployment and scalability in industrial applications.

In conclusion, this study showcases the immense potential of large language models in industrial signal processing and RUL prediction, offering a new research direction for intelligent health management in modern industrial systems. Future work may further explore the application of this framework on larger-scale industrial datasets, promoting its implementation in real-world industrial scenarios. Our research not only provides new solutions for industrial equipment maintenance but also opens new possibilities for the cross-domain application of large language models.